\pdfoutput=1
\documentclass{article}
\AtBeginDocument{%
  \providecommand\BibTeX{{%
    \normalfont B\kern-0.5em{\scshape i\kern-0.25em b}\kern-0.8em\TeX}}}
\usepackage{ijcai22}

\usepackage{times}
\usepackage{soul}
\usepackage{url}
\usepackage[hidelinks]{hyperref}
\usepackage[utf8]{inputenc}
\usepackage[small]{caption}
\usepackage{graphicx}
\usepackage{amsmath}
\usepackage{amsthm}
\usepackage{booktabs}
\usepackage{algorithm}
\usepackage{algorithmic}
\usepackage{amssymb}
\usepackage{subcaption}
\title{Learning to Compose Diversified Prompts for Image Emotion Classification
}

\author{
Sinuo Deng\and
Lifang Wu\and
Ge Shi\dag \and
Lehao Xing\and
Meng Jian\and
Ye Xiang
\affiliations
Faculty of Information Technology, Beijing University of Technology, Beijing, China
\emails
tinkersxy@gmail.com
}
\begin{document}

\maketitle

\begin{abstract}

    Contrastive Language-Image Pre-training (CLIP) represents the latest incarnation of pre-trained vision-language models. Although CLIP has recently shown its superior power on a wide range of downstream vision-language tasks like Visual Question Answering, it is still underexplored for Image Emotion Classification (IEC). Adapting CLIP to the IEC task has three significant challenges, tremendous training objective gap between pretraining and IEC, shared suboptimal and invariant prompts for all instances. In this paper, we propose a general framework that shows how CLIP can be effectively applied to IEC. We first introduce a prompt tuning method that mimics the pretraining objective of CLIP and thus can leverage the rich image and text semantics entailed in CLIP. Then we automatically compose instance-specific prompts by conditioning them on the categories and image contents of instances, diversifying prompts and avoiding suboptimal problems. Evaluations on six widely-used affective datasets demonstrate that our proposed method outperforms the state-of-the-art methods to a large margin (i.e., up to $9.29\%$ accuracy gain on EmotionROI dataset) on IEC tasks, with only a few parameters trained. Our codes will be publicly available for research purposes.

\end{abstract}

\section{Introduction}
Image Emotion Classification aims to extract emotions evoked in images. Previous methods approach this challenging but essential task by first loading a backbone that initialized on datasets of fixed label sets (i.e., ImageNet), then designing various architectures or gating, attention mechanisms to compose discriminate emotion features. Benefiting from the powerful feature composition ability of deep learning, these methods have achieved great success~\cite{zhao2022computational}.

\begin{figure}[!t]
\centering
\includegraphics[width=0.45\textwidth]{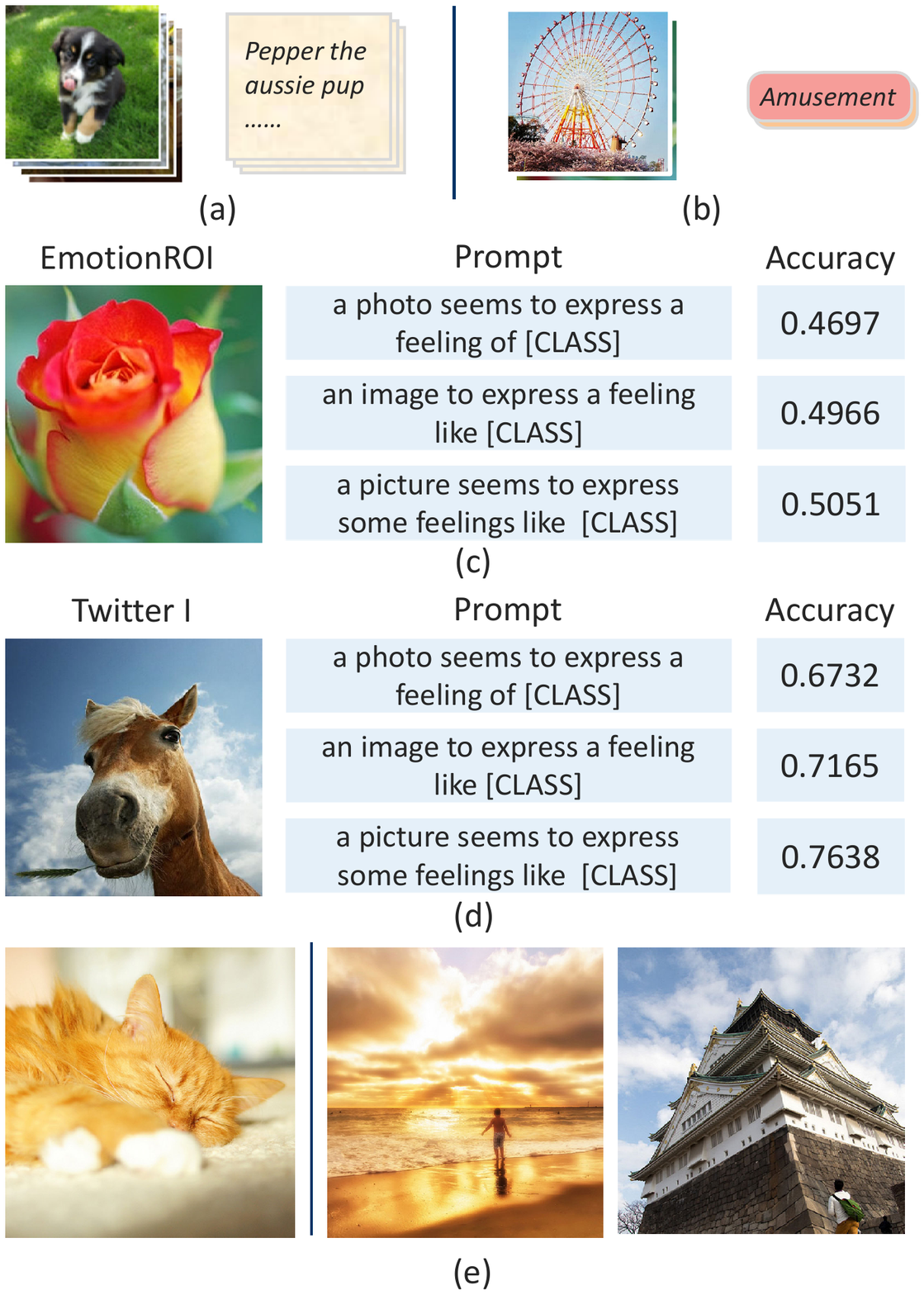} % Reduce the figure size so that it is slightly narrower than the column.
\caption{The challenges of adapting CLIP to the IEC task. (a) is the pretraining data form of CLIP while (b) is IEC task ones (i.e. FI dataset). (c), (d) show the classification results of different manually designed prompts on two affective image datasets.The left one in (e) is a picture of the \textit{contentment} category in the FI dataset, and the middle one and the right one belong to the \textit{awe} category.}
% The left one in (e) is a picture of the contentment category in the FI dataset, and the middle one and the right one belong to the awe category, while the middle one is more similar with the left one than the right one on the visual experience.}
% \Description{?}
\label{fig0}
\end{figure}

Recently, vision-language pre-training such as CLIP has emerged as a promising alternative\cite{radford2021learning}. The main idea is to align images and raw text using two separate encoders. Compared to traditional vision-only pretraining methods, the large-scale, easily accessible training data and diverse natural language descriptions enable CLIP to learn more fine-grained open-set visual concepts. Benefiting from these broader ranges of visual concepts, CLIP has strong generalization ability, and some more recent work has shown that it can be readily transferred to downstream language-vision tasks with greatly improved performance.
Adapting these recent techniques to IEC can be highly valuable, given that human emotion is highly abstractive and figuring out emotions carried in images requires a deep understanding of various details and concepts, but only a fixed set of concepts can be accessed by traditional vision pretraining methods~\cite{zhao2021affective}.

However, adapting CLIP to IEC has three significant challenges.
\textbf{(1) Tremendous gap between pretraining and IEC:} Unlike language-vision tasks, IEC has only image data during model training (as depicted in (a) (b) shown of Figure 1), which is dramatically different from CLIP training scenarios, making it difficult to effectively utilize rich knowledge entailed in CLIP.
\textbf{(2) Suboptimal prompts:} CLIP proposed to manually design text prompts to transfer knowledge to downstream tasks. However, we observe that a slight change in wording can make a massive difference in performance, as illustrated in (c) (d) of Figure 1.
\textbf{(3) Shared invariant prompts:} Some work in NLP treat prompts as sequences of virtual tokens\cite{li2021prefix} and learns prompts automatically by parameterizing these tokens\cite{lester2021power}, avoiding the suboptimal problem to some extent\cite{liu2021pre}. However, these methods employed  shared prompts across all instances, regardless of the fact that instances of different categories share similar features while also have their own distinct characteristics. As shown in (e) of Figure 1, although the left picture and the middle one are from different categories, there are some associations between them. i.e., the color. While the middle and the right ones come from one category, they have their own peculiarities.

To tackle the above three challenges, we propose a novel Prompt Tuning method with Diversified Prompt Composition (PT-DPC) based on CLIP, which can learn to compose unique prompts for each image. Specifically, we treat prompts as a sequence of tunable virtual tokens and obtain text representations by inputting them to the text encoder of CLIP, these virtual tokens are trained end-to-end and can condense the signal from a full labeled dataset, with CLIP weights fixed. We further condition these virtual tokens on the classes and image contents of instances. More specifically, we employ different virtual tokens for each class to obtain class-specific prompts. Then we integrate image contents with all class-specific prompts to compose diversified prompt, forming an instance-specific prompt and capturing associations between possible classes.  

We evaluate our model on six widely used image emotion classification benchmarks, namely, FI\_8, EmotionROI\_6, EmotionROI\_2, FI\_6, FI\_2, Twitter I, Twitter II. Experimental results show that our model outperforming state-of-the-art methods to a large margin. For example, our method achieves 9.29\% absolute accuracy gain on EmotionROI\_6 dataset. 

The contributions of this work are summarized as follows: 
\begin{itemize}
    \item We propose a novel prompt tuning method, PT-DPC, addressing three challenges of adapting the CLIP model to the IEC task. To our best knowledge, this is the first work to introduce a prompt tuning method for image emotion classification tasks.
    % affective image understanding via a Large-scaled Language-supervised Pre-training Model (LPM). To our best knowledge, this is the first work to introduce a continues prompt-tuning method for image emotion classification tasks.
    \item To avoid suboptimal problems for the fixed prompt tuning of CLIP, we propose a diversified prompt composition by utilizing both image contents and all class-specific virtual tokens.
    \item The experimental results on six popular affective image datasets demonstrate that our proposed framework can outperform the state-of-the-art methods for emotion classification.
\end{itemize}

\section{Related Work}
In this section, we will introduce the related works, including emotion representation models, image emotion classification, large-scale pre-trained models, and prompt tuning methods in NLP.
% Our work is closely related to image emotion classification and prompt tuning methods.
% description
% emotion space, emotion models: DES / CES, plutchik / mikels / ekman / polarity
% image emotion classification: traditional, cnn, architacture
% pretraining model: BERT, GPT.. CLIP
% prompt tuning: CoOp / softprompt
% 预计1.5p

\subsection{Emotion Representation Models}
In psychology theory, emotion is mainly measured by two representative models: dimensional emotion space (DES) and categorical emotion states (CES). 

DES models are designed to represent emotions by employing continuous 2D, 3D, or higher dimensional Cartesian space, such as valence-arousal (VA)~\cite{hanjalic2006extracting} and valence-arousal-dominance (VAD)~\cite{gunes2013categorical}. VAD is the most popular DES model, where valence represents the pleasantness ranging from negative to positive, arousal represents the intensity of emotion ranging from calm to excited, and the dominance represents the degree of control ranging from in control to controlled. In practice, dominance is difficult to measure and is often omitted, leading to the commonly used two-dimensional VA space. Theoretically, every emotion can be represented as a coordinate point in the Cartesian space. ~\cite{zhao2016predicting,balouchian2019lucfer}propose their researches based on this representation model, which made impressive contributions.
However, the absolute continuous values are difficult for users to distinguish, which constrains the employment of DES models.

On the contrary, although having limited emotion categories cannot well reflect the complexity and subtlety of emotions, CES models are easy for users to understand, explain, and annotate~\cite{zhao2021affective}. CES models classify emotions into a few basic categories. The simplest CES model is the sentimental binary model, which just includes negative and positive. In many cases, "emotion" is often called "sentiment", which sometimes also includes neutral. Since the sentiment is too coarse-grained, some relatively fine-grained emotion models are propose, such as Mikel's emotion wheel model(amusement, anger, awe, contentment, disgust, excitement, fear, and sadness)~\cite{mikels2005emotional} and Ekman's emotion model(anger, disgust, fear, happiness, sadness, surprise)~\cite{ekman1992argument}. In this paper, we propose an image emotion classification method that represents emotion using the CES model.

\subsection{Image Emotion Classification}
Image emotion classification is usually formulated as an emotion feature extraction problem, which represents the emotion of an image in a CES model. Learning discriminative emotion features will facilitate classification performance. To approach this task, ~\cite{machajdik2010affective,zhao2014exploring} designed many types of hand-crafted representations to bridge the affective gap between low-level features and abstract emotions in the earlier years. 

With the blooming of the convolutional neural networks (CNNs) on different tasks, researchers mainly designed CNN-based architectures~\cite{krizhevsky2012imagenet,he2016deep} that pretrained on a fixed set of labels were proposed to boost classification performance~\cite{you2015robust,yang2018weakly,deng2021emotion}.

In light of the image emotion's abstract, it is not easy to obtain sufficient discriminative features from the image itself. To boost classification performance, few efforts turn to enriching feature representations by incorporating external knowledge, such as proposing a well-designed sentiment dictionary~\cite{borth2013large,chen2014deepsentibank,wu2021discovering} or introducing different kinds of dataset-specific information~\cite{yang2017joint,rao2019multi,zhang2020weakly}.

% All the above approaches employ a fixed set of labels as supervised signals to learn visual concepts. In contrast, we use CLIP, which employs language as supervised signals, to learn more fine-grained visual concepts.

\subsection{Large-scale Pre-trained Models}
In recent years, deep neural networks, such as CNNs and attention neural networks, have been widely applied for various artificial intelligence tasks~\cite{jaderberg2015spatial,wang2017residual}. The neural models can automatically learn representations from data, thereby getting rid of complex feature engineering. However, they are easy to overfit and have poor generalization ability when lack of sufficient training data~\cite{xu2020neural}. Moreover, it is expensive and time-consuming to manually annotate large-scale data for complex tasks. Thus, it has been a critical research issue on how to train effective deep neural models for specific tasks with limited human-annotated data~\cite{HAN2021225}.

To tackle these problems, massive efforts have been devoted to manually constructing high-quality datasets~\cite{deng2009imagenet,bojar2014findings}, which triggers a wave of transfer learning~\cite{pan2009survey,thrun2012learning}.
The transfer learning formalizes a two-phase learning framework: a pre-training phase to capture knowledge from one or more source tasks, and a fine-tuning stage to transfer the captured knowledge to target tasks. Owing to the wealth of knowledge obtained in the pre-training phase, the fine-tuning phase can enable models to well handle target tasks with limited samples~\cite{HAN2021225}.

Plenty works of exploring pre-trained models (PTMs) are applied to artificial intelligence research, especially in CV and NLP.
A series of CNNs are successful used for almost all CV task, such as image classification~\cite{he2016deep} and object detection~\cite{redmon2016you}. The researchers from the NLP community develop more deep PTMs, proposing a series of powerful transformers architecture to capture the semantic meaning of the text, such as BERT~\cite{devlin2018bert} and GPT~\cite{brown2020language}.
% Training-Dataset

% %cnn训练需要更大量的数据和更高质量的标注，在一个训好的模型上调整，将会获得更好的效果。
% % imagenet上，初始构建的数据集，在很多下游task中依然可以起到很好的作用，正是因为他学习到了物体的特征获取能力。。。

% ImageNet~\cite{deng2009imagenet}

% large-scale:
% %然而在海量数据和大规模模型的今天，比imagenet规模更大的预训练模型层出不穷，很多模型比如x。x。x。。。

% multimodal
% With the application of large-scale pre-trained language models such as BERT~\cite{devlin2018bert} and GPT~\cite{brown2020language}, 

% % 数据复杂化不止是在量上，更是在维度上。多模态数据发展的今天，多模态的预训练模型也在发展。。。

% CLIP~\cite{radford2021learning}

\subsection{Prompt Tuning Method in NLP}
Nowadays, many downstream applications have achieved significant improvements by finetuning on top of the pre-trained models. However, due to the large scale of model parameters, finetuning brings a large computational and storage burden. Instead of finetuning on the full model, Prompt tuning methods transfer knowledge entailed in a large pre-trained model by designing a textual prompt to reformulate downstream tasks to look more like pretraining tasks. This reduces the gap between pretraining and downstream tasks, making the knowledge entailed in pretraining models can be transferred to downstream tasks easily with only a few annotation examples. Therefore, how to get text prompts becomes a vital issue. Currently, prompt tuning methods can be roughly divided into two categories: manually crafted and automatically learned. While manually crafting prompts~\cite{brown2020language,radford2021learning}is intuitive, creating and experimenting with these prompts takes time and experience, even experienced prompt designers may fail to manually discover optimal prompts~\cite{jiang2020can}. To automate prompt engineering, ~\cite{li2021prefix,lester2021power,zhou2021learning} parameterized the prompts by treating prompts as virtual tokens and performing prompting directly in the embedding space.  

\begin{figure*}[!ht]
\centering
\includegraphics[width=0.9\textwidth]{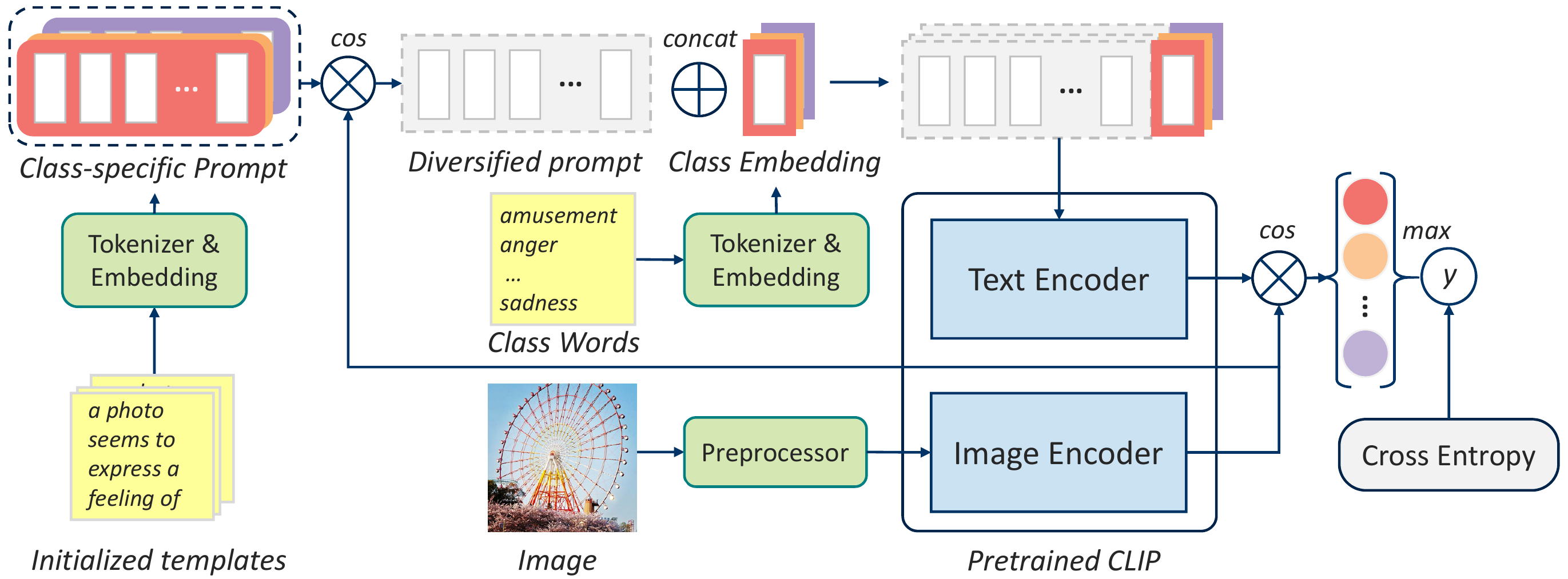} % Reduce the figure size so that it is slightly narrower than the column.
\caption{Illustration of the PT-DPC framework. 
Given an initialized template and the class words of the dataset, we generate the class-specific prompts and the class embeddings by using the same text preprocessor that includes a text tokenizer and an embedding lookup table. 
The input image is processed by the image preprocessor and image encoder to generate image features.
Then the diversified prompt is obtained by attentional filtering of the class-specific prompt with image features as queries.
After concatenating the diversified prompt with each class embedding, full diversified prompts are sent to the text encoder to get text features. 
Then the similarity scores are obtained by calculating the cosine distance between text features and image features.
We use cross-entropy loss to estimate the parameters. During model training, both the text and image encoders are frozen and only the diversified prompts are optimized.
Finally, the category to which the maximum score belongs becomes the predicted result. 
}
% \Description{?}
\label{fig1}
\end{figure*}

Our proposed model lies in the second line of work. Instead of using a shared prompt for all instances, we condition prompts on classes and contents of instances, diversifying the prompts, which can model associations between possible classes and fits pretraining scenario better.

\section{Problem Definition}
A CLIP model carries two separate encoders, image encoder $\mathcal{M}_{img}$ and text encoder $\mathcal{M}_{txt}$. And two preprocessores of image and text inputs are $E_{I}$ and $E_{T}$ respectively.
Normally, when a CLIP model is deployed on the image classification task, it is given an image $x$ with its corresponding label $y\in Y$ as input, where $Y=\{y^{(1)}, y^{(2)},...,y^{(C)}\}$ includes all $C$ categories of the dataset to which $x$ belongs. 
$f_{img}(E_{I}(x))$ is the feature representation of $x$ which obtained by $\mathcal{M}_{img}$. And $f_{txt}^{(i)}([E_{T}(p_{t});E_{T}(y^{(i)})])$ $(i\in [1, C])$ are obtained by $\mathcal{M}_{txt}$, where $p_{t}$ is a prompt that formulated by concatenating a series tokens and a class embedding $E_{T}(y^{(i)})$. $S$ means the computation of cosine similarity as Eq.~\ref{eq2}. 
\begin{equation}
S(\mathbf{a}, \mathbf{b})=\frac{\mathbf{a} \cdot \mathbf{b}}{\|\mathbf{a}\|\|\mathbf{b}\|},
\label{eq2}
\end{equation}
Then the classification result $y_{pred}$ is formed as Eq.~\ref{eq3}.
\begin{equation}
\begin{split}
y_{pred}\\
&=\arg\max_{i}S(f_{img}(E_{I}(x)),f_{txt}([E_{T}(p_{t});E_{T}(y^{(i)})]))
\\
&=\arg\max_{i}\frac{f_{img}(E_{I}(x)) \cdot f_{txt}([E_{T}(p_{t});E_{T}(y^{(i)})])}{\|f_{img}(E_{I}(x))\|\|f_{txt}([E_{T}(p_{t});E_{T}(y^{(i)})])\|}.
\end{split}
\label{eq3}
\end{equation}

We aim to find a text prompt $p_{t}$ to maximize the likelihood of $P(y_{pred}=y | p_{t})$. 

Generally, $p_{t}$ is manually crafted, which may cause the suboptimal problem. Inspired by~\cite{li2021prefix}, we parameterize $p_{t}$ by $\theta$ that can be updated, which avoiding the suboptimal problem.

\section{Methodology}
Figure~\ref{fig1} shows the overview of our PT-DPC framework. We will introduce it from two aspects: diversified prompt composition, and training.

\begin{figure}[!t]
\centering
\includegraphics[width=0.45\textwidth]{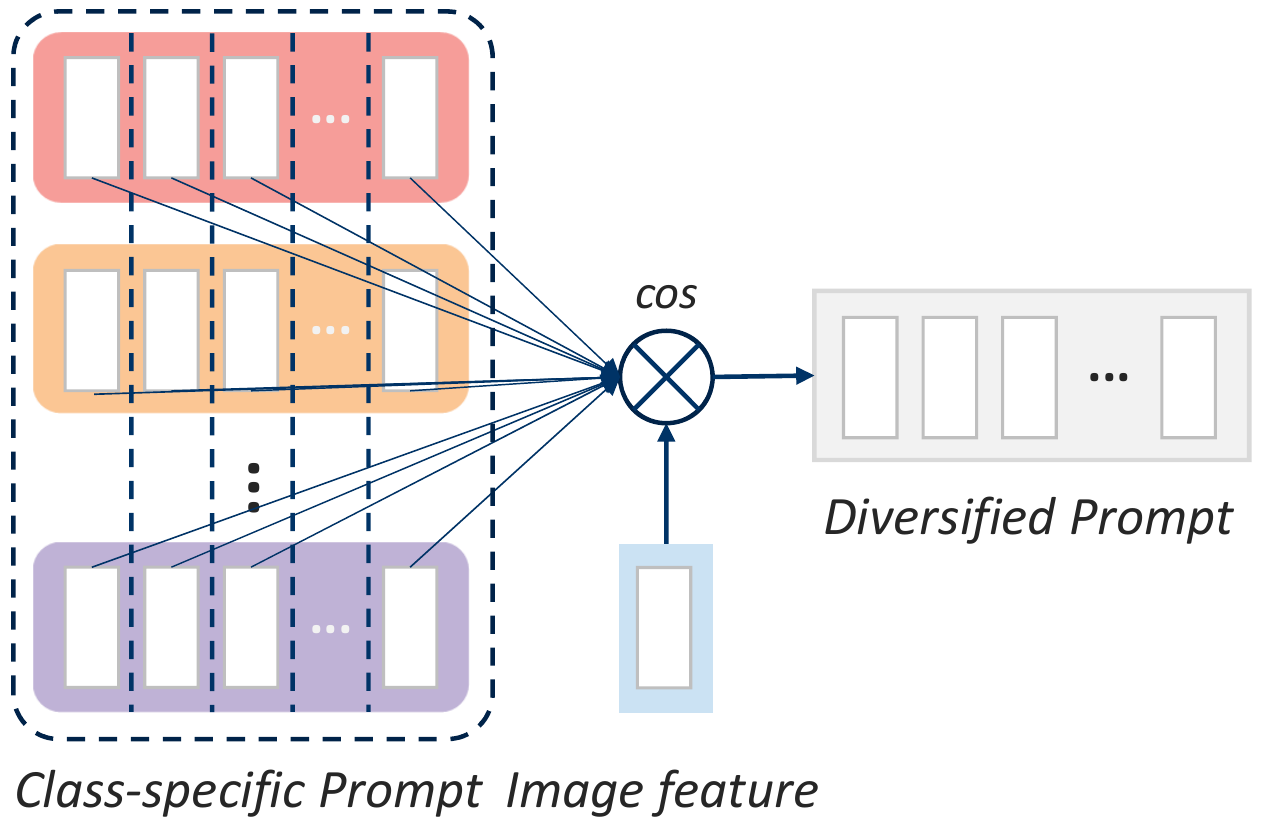} % Reduce the figure size so that it is slightly narrower than the column.
\caption{Diversified prompts construction using image features encoded by CLIP image encoder}
% \Description{?}
\label{fig2}
\end{figure}

\subsection{Diversified Prompt Composition}
Following~\cite{lester2021power}, we initialize the class-specific prompts $p_{c}^{(i)}$ by a template $s$, which are processed by the tokenizer and embedding of the original CLIP model. Then the task converts into training the parametes of $p_{c}^{(i)}$ to maximize the likelihood of $P(y_{pred}=y | p_{c})$. 
However, just like the part (e) in the Figure~\ref{fig0}, there are significant differences in the content of different affective images even though they belong to the same class. So the prompts should not be the same for different images. Thus we propose a diversified prompt composition method to utilize both the instance-specific and class-specific feature, which is shown in Figure~\ref{fig2}.

The initial parameters are the same for different categories. The effect of different initialization methods will be discussed in Section 4.4.
After the initialization of class-specific prompt $p_{c}^{(i)}$, we get $C$ series of $L$-long class embeddings which are marked as $p_{c}^{(i)(j)}$ $(j\in [1, L])$ . Each token has same dimension with the input and the output feature of the CLIP model.
We calculate the cosine similarity between the output feature of the image and each virtual token. And to multiply each virtual token which uses the similarity score as weight. Then we get the diversified prompt $p_{d}$, as the Eq.~\ref{eq1}.

\begin{equation}
\begin{split}
p_{d}(j) &= \sum_{i=1}^{C} S(f_{img}(E_{I}(x)), p_{c}^{(i)(j)}) p_{c}^{(i)(j)} \\
&= \sum_{i=1}^{C} \frac{f_{img}(E_{I}(x)) \cdot p_{c}^{(i)(j)}}{\|f_{img}(E_{I}(x))\|\|p_{c}^{(i)(j)}\|} p_{c}^{(i)(j)},
%=\frac{f_{img} \cdot f^{(i,j)}_{M_{C}}}{\|f_{img}\|\|f^{(i,j)}_{M_{C}}\|},
\end{split}
\label{eq1}
\end{equation}
where $p_{d}(j)$ is the $j$-th virual token of the diversified prompt $p_{d}$.

After composing the full diversified prompt by concatenating the $p_{d}$ with each class embedding $y^{(i)}$, the inputs of text encoder are finalized.

\subsection{Training}
\begin{equation}
\begin{split}
y_{pred}\\
&=\arg\max_{i}S(f_{img}(E_{I}(x)),f_{txt}([E_{T}(p_{d});E_{T}(y^{(i)})]))
\\
&=\arg\max_{i}\frac{f_{img}(E_{I}(x)) \cdot f_{txt}([E_{T}(p_{d});E_{T}(y^{(i)})])}{\|f_{img}(E_{I}(x))\|\|f_{txt}([E_{T}(p_{d});E_{T}(y^{(i)})])\|}.
\end{split}
\label{eq5}
\end{equation}

The final prediction $P(y_{pred})$ is obtained as Eq.~\ref{eq5}.
The entire model can be trained by maximizing the likelihood of $P(y_{pred}=y | p_{d})$ via backpropagation, while the parameters of the whole original CLIP model are fixed, gradients are only applied to update $p_{d}$.
% Cross-entropy loss is widely used in classification problems due to its advantages of fast convergence and not falling into local optimum solutions. 
We adopt cross-entropy loss as classification loss to update and optimize the model, as Eq.~\ref{eq4}.
\begin{equation}
Loss=-\frac{1}{N} \sum_{i=1}^{N} y^{(i)}\log \frac{e^{y^{(i)}}}{\sum_{j=1}^{C}e^{y^{(j)}}},
% Loss=-\sum y\log(y_{pred})
\label{eq4}
\end{equation}

\section{Experiments and Analysis}
\subsection{Datasets}
We perform experiments on four datasets with six settings, including Flickr and Instagram(FI), EmotionROI, FI\_2, EmotionROI\_2, Twitter I, and Twitter II. Following previous studies~\cite{yang2018weakly}, we adopt accuracy as the metric to evaluate our proposed method and use the same dataset split for fair comparisons.

\begin{table*}[!ht]%\scriptsize
\caption{The Classification Accuracy of PT-DPC on different datasets comparing with baseline methods}
\begin{center}
    \begin{tabular}{c|cc|cccc}
    \toprule
    Method & FI\_8 & EmotionROI\_6 & FI\_2 & EmotionROI\_2 & Twitter I & Twitter II \\ \midrule
    Zhao et al.~\cite{zhao2014exploring} & 0.4613 & 0.3484 & 0.5842     & 0.7345     & 0.6792 & 0.6751 \\ %\hline
    SentiBank~\cite{borth2013large} & 0.4923 & 0.3524 & 0.5647 & 0.6618 & 0.6663 & 0.6593 \\ \midrule
    AlexNet~\cite{krizhevsky2012imagenet} & 0.5813 & 0.4141 & 0.6863 & 0.7160 & 0.7324 & 0.7566 \\
    VGG16~\cite{simonyan2014very} & 0.6375 & 0.4546 & 0.7064 & 0.7225 & 0.7675 & 0.7699 \\
    ResNet101~\cite{he2016deep} & 0.6616 & 0.5160 & 0.7576 & 0.7392 & 0.7813 & 0.7823 \\ \midrule
    DeepSentiBank~\cite{chen2014deepsentibank} & 0.5154 & 0.4253 & 0.6154 & 0.7011 & 0.7125 & 0.7023 \\
    PCNN~\cite{you2015robust} & 0.5616 & - & 0.7534 & 0.7358 & 0.8254 & 0.7768 \\
    MldrNet~\cite{rao2020learning}  & 0.6775 & - & 0.7954 & 0.7899 & - & - \\
    Zhu et al.~\cite{zhu2017dependency} & 0.7303 & - & 0.8426 & 0.8052     & - & - \\ %\hline
    Yang et al.~\cite{yang2018visual} & - & - & 0.8635 & 0.8126 & 0.8865 & 0.8048 \\
    WSCNet~\cite{yang2018weakly} & 0.7007 & 0.5825 & -     & -     & 0.8425 & 0.8135 \\ %\hline
    ECWA~\cite{deng2021emotion} & 0.7087     & 0.5909     & - & - & 0.8479 & 0.8167 \\ %\hline
    MSRCA~\cite{zhang2022image} & 0.7260     & 0.5560     & 0.8740 & 0.8300 & - & - \\ %\hline
    Rao et al.~\cite{rao2019multi} & \textbf{0.7546} & -     & 0.8751 & 0.8294 & -     & - \\ %\hline
    Zhang et al.~\cite{zhang2020weakly} & 0.7271  & \textbf{0.6041}  & \textbf{0.9097} & \textbf{0.8510} & -     & - \\ %\hline
    Wu et al.~\cite{wu2021discovering} & -     & -     & 0.8871 & 0.8429 & \textbf{0.8965} & \textbf{0.8268} \\ \midrule
    PT-DPC & \textbf{0.7807} & \textbf{0.6970} & \textbf{0.9389} & \textbf{0.8855} & \textbf{0.9094} & \textbf{0.8250} \\ \bottomrule
    \end{tabular}%
\end{center}
\label{table2}
\end{table*}

\begin{itemize}
\item FI~\cite{you2016building} consists of 23,308 pictures distributed in 8 imbalanced emotion categories from two popular social media platforms, Flickr and Instagram. Since these eight categories can be divided into two categories according to polarity such as positive (amusement, awe, contentment, excitement) and negative (anger, disgust, fear, sadness), we also conduct experiments on binary classification setting to facilitate the comparison with work that only performs binary classification, which is called FI\_2.

\item EmotionROI~\cite{peng2016emotions} has six balanced categories of affective images, which contain 1980 images from Flickr. Since these six categories can also be divided into two categories (positive: joy, surprise, negative: anger, disgust, fear, sadness), we also evaluate our method on both a six-category setting and a binary setting, which is called EmotionROI\_2.

\item Twitter I~\cite{you2015robust} is a dataset with two categories. It has 1,269 images in total, which are collected from the Twitter platform.

\item Twitter II~\cite{borth2013large} is a small-scale dataset that contains 603 images of two different categories.
\end{itemize}

\subsection{Implementation Details}  % add more details 0.25p
We build our framework based on the CLIP-ViT-B/32~\cite{radford2021learning}, which is trained on 400 million image-text pairs and reports impressive performance on several zero-shot downstream tasks. The CLIP model has been pre-trained on a large-scale dataset, and the CLIP text and image encoders are fixed throughout the experimental period.

We employ the SGD optimizer to tune the trainable part with $0.1$, $0.01$, and $0.001$ as the initial learning rate of different datasets with $0.9$ momentum. And the learning rate was updated by a StepLR scheduler which stepsize is $3$ and gamma is $0.9$.

All our experiments are carried out on an NVIDIA RTX3090 GPU with 32GB of CPU memory using PyTorch framework\cite{paszke2019pytorch}. Images are resized and center cropped to $224\times224$, channel converted, and data normalized by the original CLIP project's preprocessor, with a batch size of $64$ for $10$ epochs.

\subsection{Classification Performance}
In this section, we review recent works on image emotion classification and compare them with our method. There are two hand-craft-feature-based methods, three CNN-based finetuning methods, and a few unique pipeline design methods based on the CNN backbone in recent years, including the SOTA methods of each dataset.

\begin{table*}[!ht]%\scriptsize
\caption{Ablation Study Results of PT-DPC on Different Datasets}
\begin{center}
    \begin{tabular}{cc|cc|cccc}
    \toprule
    IS & CS & FI\_8 & EmotionROI\_6 & FI\_2 & EmotionROI\_2 & Twitter I & Twitter II \\ \midrule
    &  & 0.7730 & 0.6380 & 0.9372 & 0.8771 & 0.8937 & 0.8083 \\ 
    \checkmark &  & 0.7792 & 0.6481 & 0.9372 & 0.8788 & 0.8976 & 0.7667 \\ %\hline
    & \checkmark  & 0.7760 & 0.6919 & 0.9381 & 0.8737 & 0.8937 & 0.8167 \\ \midrule
    \checkmark & \checkmark & \textbf{0.7807} & \textbf{0.6970} & \textbf{0.9389} & \textbf{0.8855} & \textbf{0.9094} & \textbf{0.8250} \\
     \bottomrule
    \end{tabular}%
\end{center}
\label{table3}
\end{table*}

\begin{itemize}
    \item In the early years, researchers explored emotion classification tasks in terms of hand-craft features at the image art level\cite{zhao2014exploring} or using sentiment dictionary and simple classifiers\cite{borth2013large}. 

    With the rise of the deep learning method, the image emotion classification methods turned to use CNN as a backbone, getting better performance. 
    \item DeepSentibank\cite{chen2014deepsentibank} employed CNNs to discover ANPs and realized visual sentiment concept classification. 
    \item PCNN~\cite{you2015robust} proposed a novel progressive CNN architecture based on VGGNet.
    \item Yang et al.\cite{yang2018visual} employed object detection technique to produce the “Affective Regions” and proposed three fusion strategies to generate the final predictions on VGGNet. 
    
    ResNet is also a widely-used CNN baseline structure. It is pre-trained on the ImageNet dataset~\cite{deng2009imagenet} and fine-tuned after modifying its FC layer. 
    \item Based on the backbone of ResNet101, the WSCNet~\cite{yang2018weakly} realized the end-to-end image emotion classification by coupling the global and local features according to the detected salient regions in the image, which is the best method among the content-based image emotion classification task. 
    \item To compare with the WSCNet, ECWA\cite{deng2021emotion} proposed an emotion class-wise aware loss on the same backbone. Only fine-tuning the backbone, it got better performance on all datasets than WSCNet without any other structure. 
    \item Zhu et al.\cite{zhu2017dependency} explored a unified CNN-RNN architecture for visual emotion recognition. 
    \item MldrNet\cite{rao2020learning} provided a CNN architecture based on AlexNet with a side branch to utilize hieratical features. 
    \item MSRCA\cite{zhang2022image} proposed a novel multi-level sentiment region correlation analysis model.

    \item Due to some datasets having probabilities with their corresponding labels,~\cite{rao2019multi} utilized the label probability of the affective images into a loss function for training to leverage the ambiguity and subjectivity of the class labels. This work has achieved the best classification performance on several benchmark datasets~\cite{zhao2021affective}. 
    \item~\cite{zhang2020weakly} proposed an end-to-end network for IEC leveraging weakly supervised emotion intensity learning, achieved SOTA performance on FI\_2 and two categories types of EmotionROI dataset. 
    \item Based on the object information in the detected image,~\cite{wu2021discovering} built a graph convolutional network based on the sentiment dictionary to explore the relationship among the object in the image, which made a better performance on the sentiment polarity classification datasets.
\end{itemize}

The experimental results are shown in the Table~\ref{table2}. 
Despite having better interpretability, early hand-crafted feature methods, such as Zhao et al.~\cite{zhao2014exploring} and Sentibank~\cite{borth2013large} are generally less effective than deep learning methods. Then with the rise of deep learning, the performance of deep feature methods on IEC improves with the depth of network architecture and the number of model parameters, like the AlexNet~\cite{krizhevsky2012imagenet}, VGG-16~\cite{simonyan2014very}, and ResNet101~\cite{he2016deep}. 

With the blooming development of deep learning architecture, more and more researchers managed to capture the internal sentimental factors ~\cite{you2015robust,yang2018weakly,deng2021emotion} or utilize external knowledge ~\cite{chen2014deepsentibank,yang2017joint,rao2019multi,zhang2020weakly,wu2021discovering} to improve the classification performance. They get better results than the original deep models.
Our method, PT-DPC, which utilizes a large scale pre-trained model with richer knowledge, achieves competitive results on all six commonly used datasets, e.g., our method achieves about $2.9\%$ improvement on the FI\_2 dataset and $9.29\%$ on EmotionROI\_6 than the-state-of-the-art methods. Only the TwitterII dataset gets the second place by $0.18\%$ difference. It is mainly because of the tiny data scale, with only $603$ images in total.

\begin{table*}[!ht]%\scriptsize
\caption{Sensitivity Analysis Results of PT-DPC on Different Datasets}
\label{table4}
\begin{center}
    \begin{tabular}{c|cc|cccc}
    \toprule
    Template & FI\_8 & EmotionROI\_6 & FI\_2 & EmotionROI\_2 & Twitter I & Twitter II \\ \midrule
    % PT-DPC-1 & 0.7730 & 0.7003 & 0.9357 & 0.8687 & 0.8543 & 0.8167 \\ 
    PT-DPC-1   & 0.7807 & 0.6970 & 0.9389 & 0.8855 & 0.9094 & 0.8250 \\ 
    PT-DPC-2 & 0.7807 & 0.6717 & 0.9407 & 0.8872 & 0.8898 & 0.8083 \\ 
    PT-DPC-3 & 0.7824 & 0.6751 & 0.9357 & 0.8788 & 0.8937 & 0.8167 \\ \midrule
    Std. & 0.0010 & 0.0137 & 0.0025 & 0.0044 & 0.0104 & 0.0005 \\ \bottomrule
    \end{tabular}%
\end{center}
\end{table*}

\begin{table*}[!ht]\small
\caption{The Classification Accuracy of different vision encoder on the affecitve image datasets}
\begin{center}
    \begin{tabular}{c|cc|cccc}
    \toprule
    Method & FI\_8 & EmotionROI\_6 & FI\_2 & EmotionROI\_2 & Twitter I & Twitter II \\ \midrule
    ImageNet-AlexNet~\cite{krizhevsky2012imagenet} & 0.3826 & 0.3426 & 0.6054 & 0.6465 & 0.6580 & 0.6788 \\
    ImageNet-VGG16~\cite{simonyan2014very} & 0.4122 & 0.3726 & 0.7064 & 0.7225 & 0.6749 & 0.6879 \\
    ImageNet-Res101~\cite{he2016deep} & 0.5001 & 0.4079 & 0.8142 & 0.7593 & 0.7255 & 0.7042 \\ \midrule
    % Fine-tuned AlexNet~\cite{yang2018weakly} & 0.5813 & 0.4141 & 0.6863 & 0.7160 & 0.7324 & 0.7566 \\
    % Fine-tuned VGG16~\cite{yang2018weakly} & 0.6375 & 0.4546 & 0.7064 & 0.7225 & 0.7675 & 0.7699 \\
    % Fine-tuned Res101~\cite{yang2018weakly} & 0.6616 & 0.5160 & 0.7576 & 0.7392 & 0.7813 & 0.7823 \\ \midrule
    % MAE~\cite{he2021masked} & 0.3271 & 0.2677  & 0.7111 & 0.6599 & 0.6220 & 0.8000 \\
    DeiT~\cite{touvron2021training} & 0.6494 & 0.4949 & 0.8500 & 0.7609 & 0.7953 & 0.7833  \\
    BEiT~\cite{bao2021beit} & 0.7261 & 0.5993 & 0.9063 & 0.8165 & 0.8386 & 0.7667 \\ \midrule
    CLIP-ViT~\cite{radford2021learning} & 0.7623 & 0.6717 & 0.9164 & 0.8855 & 0.8701 & 0.8083 \\
    CLIP~\cite{radford2021learning} & 0.7801 & 0.6633 & 0.9349 & 0.8788 & 0.8937 & 0.8167 \\ \midrule
    PT-DPC & \textbf{0.7807} & \textbf{0.6970} & \textbf{0.9389} & \textbf{0.8855} & \textbf{0.9094} & \textbf{0.8250} \\ \bottomrule
    \end{tabular}%
\end{center}
\label{table5}
\end{table*}

\subsection{Ablation Study} 
The diversified prompt part utilizes the instance-specific information and the class-specific information to leverage the knowledge from the large-scale pre-trained model.
To evaluate the effectiveness of the two proposed components, we conduct the ablation study on all six dataset. The experimental results are shown in Table~\ref{table3}.
% The full model line is the normal version of PT-DPC that we have shown above. 
% Especially, "\textit{IS}" represents instance-specific information, and "\textit{CS}" represents class-specific information. 

\textbf{Effectiveness of the instance-specific information.} The first row in Table~\ref{table3} denotes the baseline single prefix tuning method like~\cite{lester2021power}, which doesn't use any instance-specific information or class-specific information. The second row means we adopt the instance-specific component, which means the prompt templates are initialized separately. And the diversified prompt is obtained by utilizing its corresponding image as query. The results show that utilizing the instance-specific information can moderately improve the classification performance on most datasets% , which indicate
.

\textbf{Effectiveness of the class-specific information.} We show the model performance of utilizing class-specific information in the third row in Table~\ref{table3}, while the fourth row means we further utilize class-specific information on the basis of having instance-specific information. We can see that the class-specific information can improve the model performance on all but one of the six datasets.

Totally, utilizing instance-specific information improves the accuracy while leveraging class-specific information further improves the performance, which indicates the necessity of considering both the instance-specific information and the class-specific information. Our proposed PT-DPC achieved the best performance by considering both the instance-specific and class-specific information, which indicates that both the image content and sentiment concept are important for prompt tuning on IEC task.

% When the experiment is run without "\textit{IS}", it means that the prompt is formed as the mean value of class-specific embedding without multiplying similarity weights from the corresponding image feature. Compared with the original prompt template, there is an overall improvement after adding the class-specific information.

% When ablating "\textit{CS}", the prompt template is initialized separately. And the diversified prompt is obtained by weighting the similarity of its corresponding image feature. The results show a relative improvement in different datasets, and even in individual datasets, better results are obtained than adding only category information.

% Ultimately, the best results are obtained in the whole dataset when the two parts of information are fused. This illustrates that the two components, which have validity individually, can obtain better results when used in a fusion.

\subsection{Sensitivity analysis about the Prompt Initialization}

The performance of previous prompt tuning methods are highly sensitive to prompts, so it is necessary to take the sensitivity analysis of the prompt initialization for PT-DPC.

There are different massive types of the template to initialize the prompt. Inspired by the prompt tuning work, we chose a few kinds of standing prompt templates for this ablation study.
\begin{itemize}
    \item PT-DPC-1 is a template of $"a$ $photo$ $seems$ $to$ $express$ $a$ $feeling$ $of$ $[label$ $word]"$, which is similar with the original CLIP type. We use this template for performance comparison in Section 4.3.
    
    \item PT-DPC-2 is a template of $"an$ $image$ $to$ $express$ $a$ $feeling$ $like$ $[label$ $word]"$, which some words are replaced by synonyms.
    
    \item PT-DPC-3 is a template of $"a$ $picture$ $seems$ $to$ $express$ $some$ $feelings$ $like$ $[label$ $word]"$, which are replaced in the same meaning.
\end{itemize}
Except for the classification results of each initialized template, we also calculate the Standard Deviation(Std) of these results on different datasets to intuitive display the degree of influence of different templates. The smaller the Std is, the better the PT-DPC's sensitivity on the corresponding dataset.

As shown in Table~\ref{table4}, different initialization only brings small changes and is not relevant to a particular dataset. The Std results show only a few influences on different initialization methods. It almost does not influence the FI and TwitterII dataset with polarity classification. Though further tuning of the initialization words might help, it can still consider that PT-DPC is robust by different initialized templates.

\begin{figure*}[!t]
 \centering
    %  \subfigure[(a)]
     {
      \begin{minipage}[t]{0.46\textwidth}
       \centering
       \includegraphics[width=0.98\textwidth]{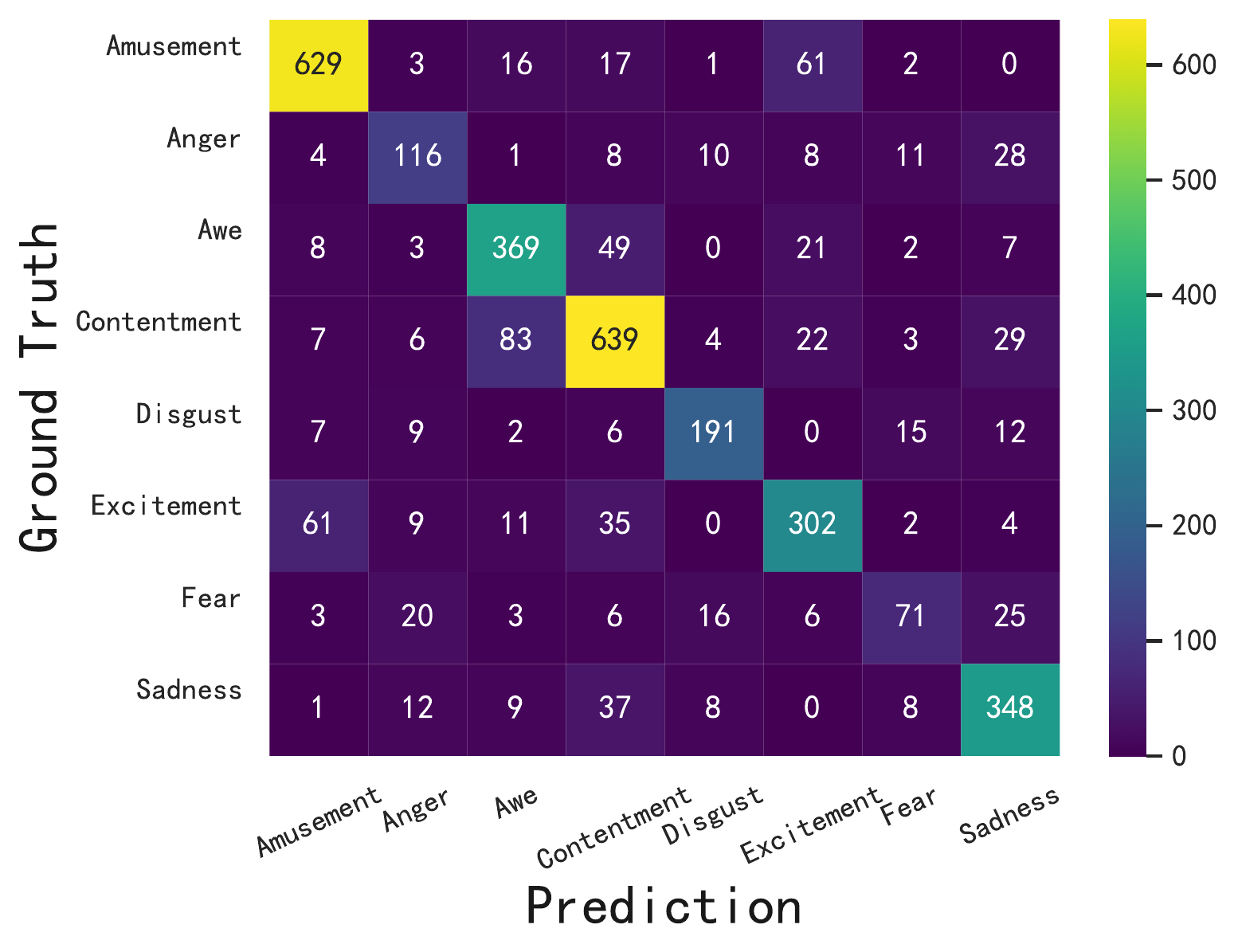}
       \subcaption{FI\_8}
      \end{minipage}
     }
    % \subfigure[(b)]
    {
     \begin{minipage}[t]{0.45\textwidth}
      \centering
      \includegraphics[width=0.96\textwidth]{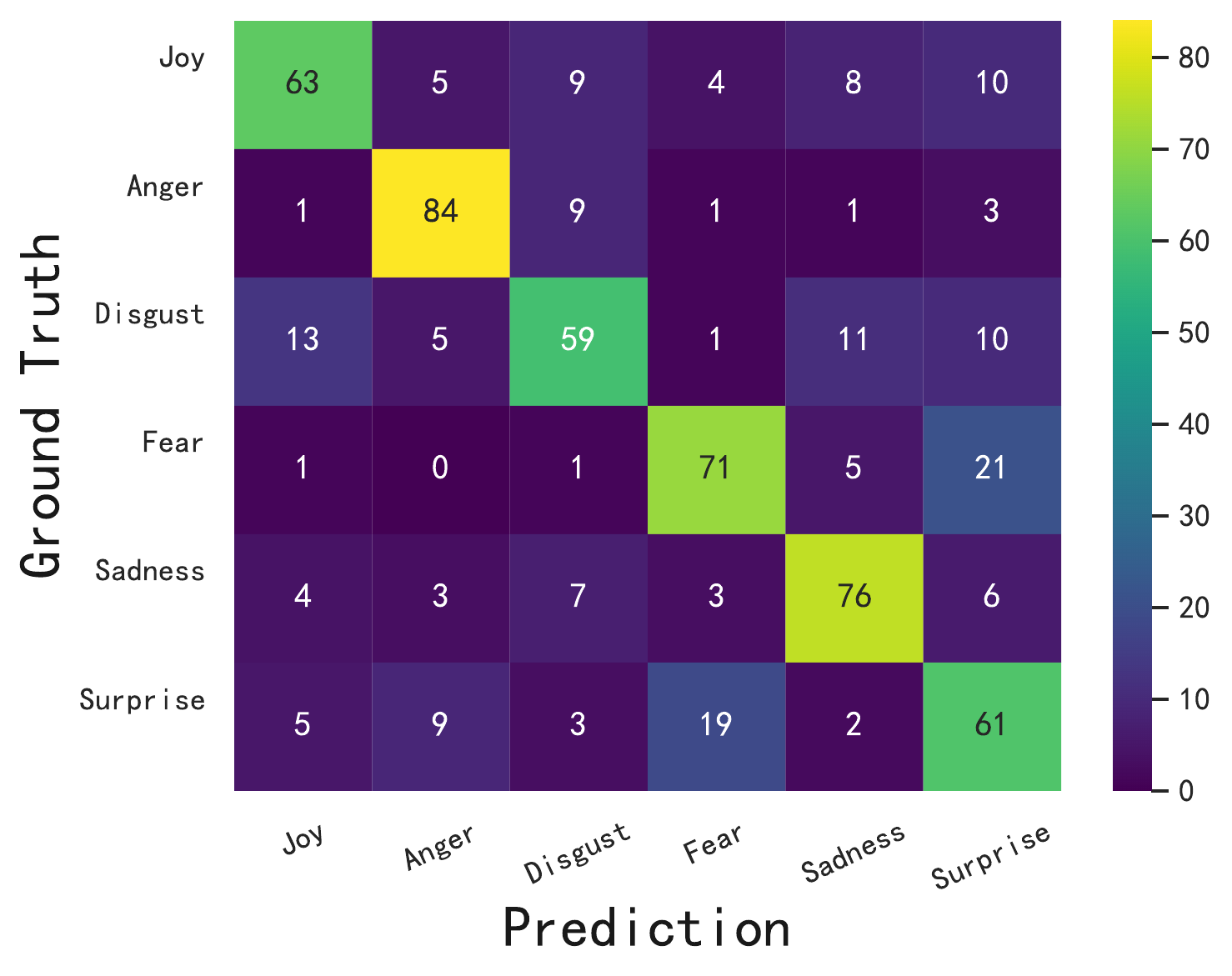}
      \subcaption{EmotionROI\_6}
     \end{minipage}
    }
\caption{Visualization confusion metric of the classification results of PT-DPC method on the FI\_8 and EmotionROI\_6 dataset. The different colors distinguish different accuracy degree as shown in color bar. (a) shows the results of FI\_8 dataset, while (b) presents the EmotionROI\_6.}
\label{figcm}
\end{figure*}

\begin{figure*}[!t]
\centering
\includegraphics[width=0.9\textwidth]{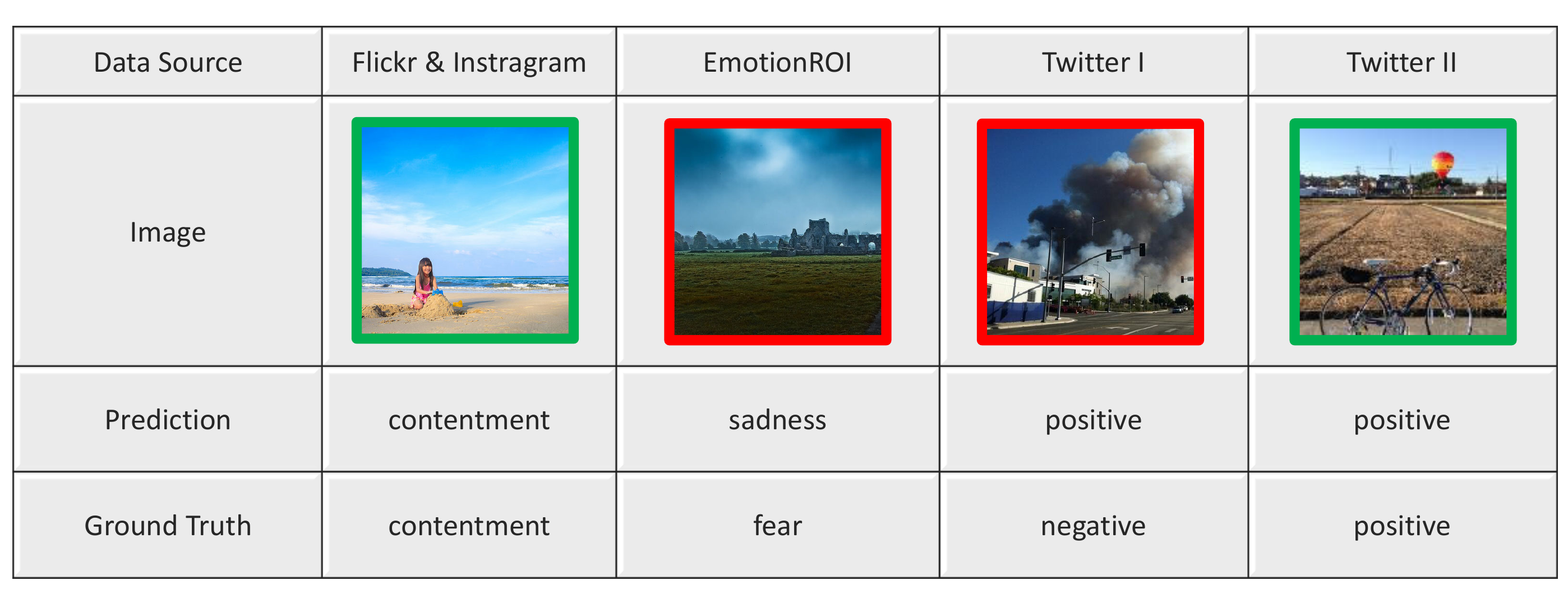} % Reduce the figure size so that it is slightly narrower than the column.
\caption{Case example of the four datasets.}
% \Description{?}
\label{figcase}
\end{figure*}

% \begin{figure}[!t]
% \centering
% \includegraphics[width=0.47\textwidth]{new8.pdf} % Reduce the figure size so that it is slightly narrower than the column.
% \caption{Visualization confusion metric of the classification results of PT-DPC method on the FI\_8 dataset. The different colors distinguish different accuracy degree as shown in color bar.}
% \Description{?}
% \label{figfi8}
% \end{figure}

% \begin{figure}[!t]
% \centering
% \includegraphics[width=0.47\textwidth]{new6.pdf} % Reduce the figure size so that it is slightly narrower than the column.
% \caption{Visualization confusion metric of the classification results of PT-DPC method on the EmotionROI\_6 dataset. The different colors distinguish different accuracy degree as shown in color bar.}
% \Description{?}
% \label{figem6}
% \end{figure}

\subsection{Impact of Vision Encoder}
From the earlier CNNs to Vision Transformers, many vision encoder architectures have been developed to capture discriminate feature representations. Thus we conduct the classification experiment among the different vision encoder architectures employed in the comparison methods. 
In detail, to achieve the classification on each dataset, we change the dimension of the output layer and only train this fully-connected layer but fix the other part of the encoder.

The experimental results are shown in the Table~\ref{table5}. The CNNs model performs better with the deeper depth the models are. 
% And as reported by Yang et al.~\cite{yang2018weakly}, it can get a better result if the model is appropriately finetuned.
Compared with CNNs, the transformer models can achieve better performance. We conduct a simple but same linear-probe experiment on two famous image encoders based on transformer architecture. And for fair comparison without text influence, besides the CLIP model, we also employ a single image encoder of CLIP, CLIP-ViT. Unsurprisingly, the CLIP-ViT outperforms each other vision encoder, and the CLIP is much better on average. 
It illustrates that pre-training or training with sufficient good semantic supervision is beneficial for the model to perform the emotion recognition task.
% \subsection{Case Analysis}

\section{Visualization}
To demonstrate the performance of PT-DPC on the image sentiment classification task more intuitively, we employ the confusion matrix in the Figure~\ref{figcm}.
The numbers in the matrix indicate the number of ground truth category images predicted into different categories in the corresponding test set. The higher the number is, the higher the corresponding grids' colour brightness is. Bright diagonal grids in the figure mean that we have excellent classification results in the two primary datasets.

% \section{Case Analysis}

\section{Remaining Challenges}
The successfully applying of large models has a strong effect on recognising semantically complex affective images. However, a unique domain adaptation design is needed for datasets with small-scale and large domain biases, such as the TwitterII dataset, to bridge this gap more adequately.
In addition, the emotion is complicated and there is not an either/or relationship between affective categories. 

As can be seen from Fig~\ref{figcase}, one image from each dataset is exhibited in a row. They are informative that different reasons cause them to be attributed to different emotional categories. However, ignoring some factual elements can lead to erroneous judgments. For example, when there is no concept of disaster, the third picture will only be misclassified as positive for being identified as a spectacular scene.
At the same time,  since some of the images could have belonged to more than one emotion category, they would have produced false results like the second image, which seems acceptable.
So it could be considered to further research about the interclass relationships or LDL-related studies by PT-DPC or to exploit a more large-scale pre-trained model beyond CLIP.

\section{Conclusion}
In this paper, we first propose a general framework that adapting CLIP to the image emotion classification task, diversified prompt composing (PT-DPC), to effectively leverage the rich image and text semantics entailed in CLIP. 
Except for addressing the challenge of the training objective gap, the PT-DPC automatically compose instance-specific prompts by conditioning them on the categories and image contents of instances, diversifying prompts and avoiding suboptimal problems. 
Compared with the state-of-the-art method, PT-DPC performs better in several widely-used datasets, including binary categories and multi-categories.
Furthermore, research about the multi-modal method is well popular, but there are still many cases alive with only one modal information. We hope that the ideas in this article can inspire other resource-constrained tasks like image emotion classification and develop more novel multi-modal methods for traditional single-modal tasks.

\section*{Acknowledgements}
This work was supported in part by the National Natural Science Foundation of China under Grant NO. 62106010, 61976010, 62176011, 62106011.

% \clearpage
\small

\bibliographystyle{named.bst}
\bibliography{arxiv.bib}

\end{document}